\documentclass[11pt,a4paper]{article}
\usepackage[hyperref]{udst2018}
\usepackage{times}
\usepackage{latexsym}
\usepackage{url}
\usepackage[utf8]{inputenc}
\usepackage{float}

\usepackage{tikz}
\usepackage{tkz-graph}
\usetikzlibrary{arrows,calc,matrix,trees,patterns,automata,positioning,shapes.geometric}

\aclfinalcopy 
\def\aclpaperid{***} 

\title{Semi-Supervised Neural System for Tagging, Parsing and Lemmatization}

\author{Piotr Rybak \\
  Institute of Computer Science, \\
  Polish Academy of Sciences \\
  {\tt piotr.cezary.rybak@gmail.com} \\\And
  Alina Wróblewska \\
  Institute of Computer Science, \\
  Polish Academy of Sciences \\
  {\tt alina@ipipan.waw.pl} \\}

\date{}

\begin{document}
\maketitle

\newcommand{\udst}[0]{\emph{CoNLL 2018 UD Shared Task}}

\begin{abstract}
This paper describes the ICS PAS system which took part in CoNLL 2018 shared task on Multilingual Parsing from Raw Text to Universal Dependencies. The~system consists of jointly trained tagger, lemmatizer, and dependency parser which are based on features extracted by a~biLSTM network. The~system uses both fully connected and dilated convolutional neural architectures. The~novelty of our approach is the~use of an~additional loss function, which reduces the~number of cycles in the~predicted dependency graphs, and the~use of self-training to increase the~system performance. The~proposed system, i.e. ICS PAS (Warszawa), ranked
3th/4th in the~official evaluation\footnote{\url{http://universaldependencies.org/conll18/results.html}} obtaining the~following overall results: 73.02 (LAS), 60.25 (MLAS) and 64.44 (BLEX).

\end{abstract}

\section{Introduction}
Most of contemporary NLP systems for machine translation,  question answering, sentiment analysis, etc. operate on preprocessed texts, i.e. texts with tokenised, part-of-speech tagged, and 
possibly syntactically parsed sentences. Therefore, the~development of high-quality pipelines of NLP tools or entire systems for language preprocessing is still an~important issue. The~vast majority of language preprocessing frameworks take advantage of the~statistical methods, especially the~supervised or semi-supervised statistical methods. Based on training data, language preprocessing tools learn to analyse sentences and to predict morphosyntactic annotations of these sentences. 

The~supervised methods require gold-standard training data whose creation is a~time-consuming and expensive process. Nevertheless, the~morphosyntactically annotated data sets are publicly available for many languages, in particular within Universal Dependencies initiative \cite[UD,][]{NivreEtAl:2016}. The~initiators of UD aim at developing a~cross-linguistically consistent  annotation schema and at building a~large multilingual collection of sentences annotated according to this schema with the~universal part-of-speech tags and the~universal dependency trees.

UD treebanks are nowadays used for multilingual system development 
\cite{ud22data}. The~history of developing multilingual systems dates back to 2006 and 2007, when two shared tasks on multilingual dependency parsing were organised at the~Conference on Computational Natural Language Learning \cite[CoNLL,][]{BuchholzAndMarsi:2006,Nivreetal:2007}. After 10 years, the~shared task was organised again in 2017 \cite{udst:overview2017}, and currently there is its fourth edition \cite{udst:overview}.

In this paper we describe our solution submitted to the~CoNLL 2018 Universal Dependency shared task. The~system and the~trained models for participating treebanks are publicly available.\footnote{\url{https://github.com/360er0/COMBO}}
Our system takes a~
tokenised sentence as input. The~sentence tokenisation is predicted by the~baseline model \cite{udpipe:2017}.

\begin{figure}[h!]
\begin{center}

\includegraphics[scale=0.75]{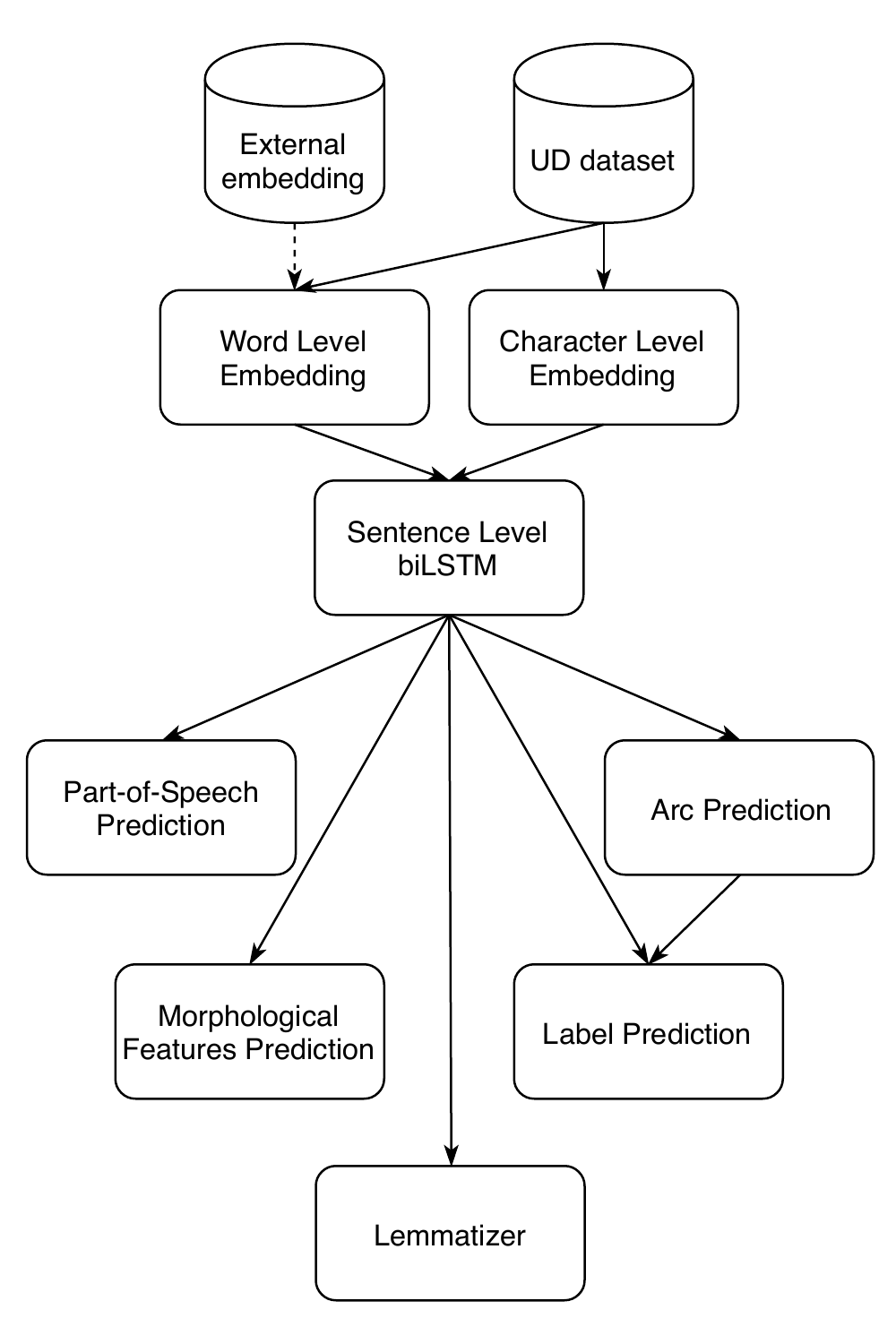}

\caption{\label{fig:overview}The~
schema of the~system architecture.}
\end{center}
\end{figure}

Each word is represented both as an~external word embedding and as a~character-based word embedding estimated by a~dilated convolutional neural network encoder \cite[CNN,][]{dcnn:2015}. The concatenation of these embeddings is fed to a~bidirectional long short-term memory network \cite[biLSTM,][]{bilstm:2005,bilstm:1997} which extracts the final features (see Section \ref{sec:featureExtraction}). The tagger takes extracted features and predicts universal part-of-speech tags, language-specific tags and morphological features using three separate fully connected neural networks with one hidden layer (see Section \ref{sec:tagger}). The~lemmatizer uses a~dilated CNN to predict lemmas based on characters of corresponding words and features previously extracted by a~biLSTM encoder (see Section \ref{sec:lemmatizer}). As a scoring function, the~graph-based dependency parser uses simple dot product of the~vector representations of a~dependent and its governor. These representations are output by two single fully connected layers which take feature vectors extracted by a~biLSTM encoder as input.
A~novel loss function penalizes cycles, in order to reduce their number in the~predicted dependency graphs (see Section \ref{sec:loss}). Chu-Liu-Edmonds algorithm \cite{chuLiu:65,Edmonds:67} constructs the~final dependency tree. The~dependency labels are predicted with a~fully connected neural network based on the~dependent and its governor embeddings as well (see Section \ref{sec:parser} for more details on the~parser's architecture). The~system architecture is schematised in Figure \ref{fig:overview}.
%
%

The~whole system is end-to-end trained, separately for each treebank provided for the~purposes of the~shared task. The~technical details of the~implemented system are given in Section \ref{sec:implementationDetails}. Additionally, for 20 selected treebanks self-training is used to increase the~performance of the~models (see Section \ref{sec:selflearning}). The~proposed technique of self-training has an~impact on the~quality of tagging, lemmatisation and parsing (see Section \ref{sec:results1}). The~article ends with the~presentation of the~results achieved by our system (see Section \ref{sec:resultOverview}) and some conclusions (see Section \ref{sec:conclusions}).

\section{Architecture Overview}
\subsection{Feature Extraction}
\label{sec:featureExtraction}
The~system accepts an~input in the~form of tokenised sentences that can be annotated with additional morphosyntactic information: lemmas, part-of-speech tags, and morphological features. However, as the~goal of the~shared task is to predict not only dependency trees but also parts of speech, lemmas and morphological features,\footnote{Not all treebanks are annotated with lemmas and morphological features.} we decide to use words as the~only input. 

\subsubsection{Word Level Embedding}
Each input word is represented as a~vector using the~external pre-trained embedding. Words not present in the~external embedding are replaced with the~``unknown'' word and represented as a~random vector drawn from the~normal distribution with the~mean and the~variance calculated based on other word embedding vectors. Both the~external embedding itself and the vector representing ``unknown'' word are fixed during the~training, but they are transformed by a~single fully connected layer. This transformation serves similar purpose as a~trainable embedding, but helps with generalization, since it will also transform vectors for words available in the~external embedding, but not in the~training set. 

\subsubsection{Character Level Embedding}
Additionally, each word is represented as the~character-based word embedding extracted with a~dilated convolutional neural network (CNN). We decide to use the~dilated CNN instead of 
commonly used biLSTM encoder to speed up the training of the system.

First, each word is transformed to a~sequence of the~trainable character embeddings. Moreover, the~special symbols ``beginning-of-word'' and ``end-of-word'' are added to the~sequence to represent the~beginning and the~end of the~word. Then the~dilated CNN encoder is used. Since the~encoder also outputs a~sequence, we use the~global max-pooling operation to obtain the~final word embedding. This procedure is reasonable for estimating embeddings of out-of-vocabulary words, especially in languages with rich morphology.

\subsubsection{Sentence Level biLSTM}
Both word representations are concatenated together and fed into the~sentence level biLSTM network. 
The~network learns contexts for each 
word and extracts the~final features for each of these 
words.

\subsection{Tagger}
\label{sec:tagger}
\subsubsection{Part-of-Speech Tags}
The~tagger is implemented as a~fully connected network with one hidden layer and soft-max activation function. The~tagger takes the~features extracted by the~biLSTM as input and predicts a~universal part-of-speech tag and a~language-specific tag for each 
word.

\subsubsection{Morphological Features}
\begin{figure}[H]
\begin{center}

\includegraphics[scale=0.75]{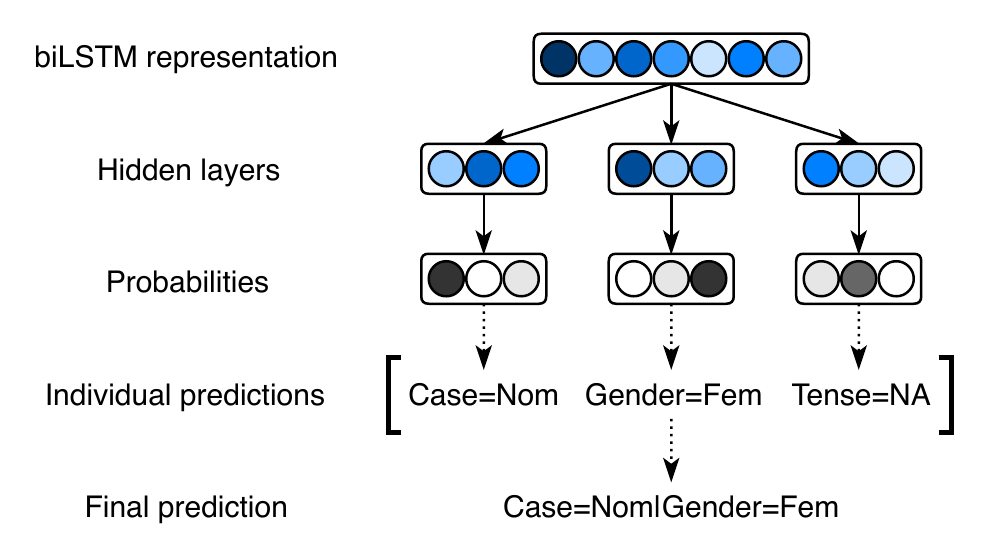}
\caption{\label{fig:morf}The morphological features prediction.}
\end{center}
\end{figure}

Similar approach is used to predict morphological features. 
Each morphological feature is represented as an~attribute-value pair (e.g. Number=Sing) and each word is annotated with a~set of appropriate attribute-value pairs in training data. We therefore decide to treat the~problem of morphological features prediction as several classification problems (see Figure \ref{fig:morf}).

For each attribute its value is predicted with a~fully connected network with one hidden layer and soft-max activation function. Various 
words are defined by the~sets of various morphological features. Since for each 
word only some attributes are present in the~set of  morphological features, the possible values are extended with ``not 
applicable'' label. It allows the~model to learn that an~attribute is not present in the~set of morphological features of a~particular word.
%
%

\subsection{Lemmatizer}
\label{sec:lemmatizer}
Lemmatizer takes two different inputs. First, features extracted by the~biLSTM encoder are used, however their dimensionality is reduced with a~single fully connected layer. Next, the~word, for which we want to predict a~lemma, is converted to a~sequence of characters. The special symbols ``beginning-of-word'' and ``end-of-word'' are added to the sequence to represent the~beginning and the~end of the~word. Each character in the~sequence is represented as a~trainable embedding vector. The~final input to the~lemmatizer is a~sequence of character embeddings concatenated with the~reduced version of features extracted by the~biLSTM encoder. Note that each character embedding is concatenated with exactly the~same extracted feature vector.

\begin{figure}[H]
\begin{center}

\includegraphics[scale=0.75]{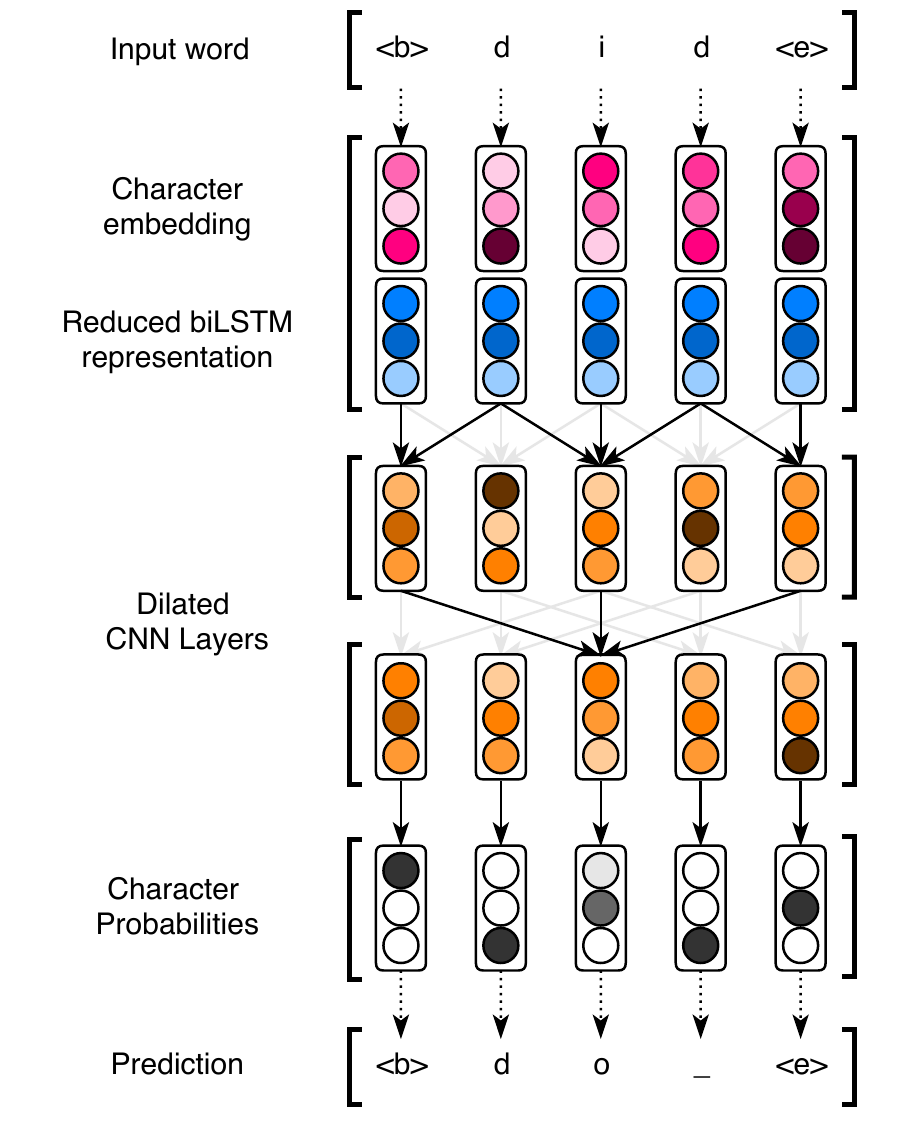}
\caption{\label{fig:lemma}The~lemmas prediction.}
\end{center}
\end{figure}

Then the~dilated convolutional neural network followed by soft-max function converts given input to the~sequence of probabilities of one-hot encoded characters of the~predicted lemma (see Figure \ref{fig:lemma}).

\subsection{Parser}
\label{sec:parser}
\subsubsection{Arc Prediction}
\begin{figure}[H]
\begin{center}

\includegraphics[scale=0.75]{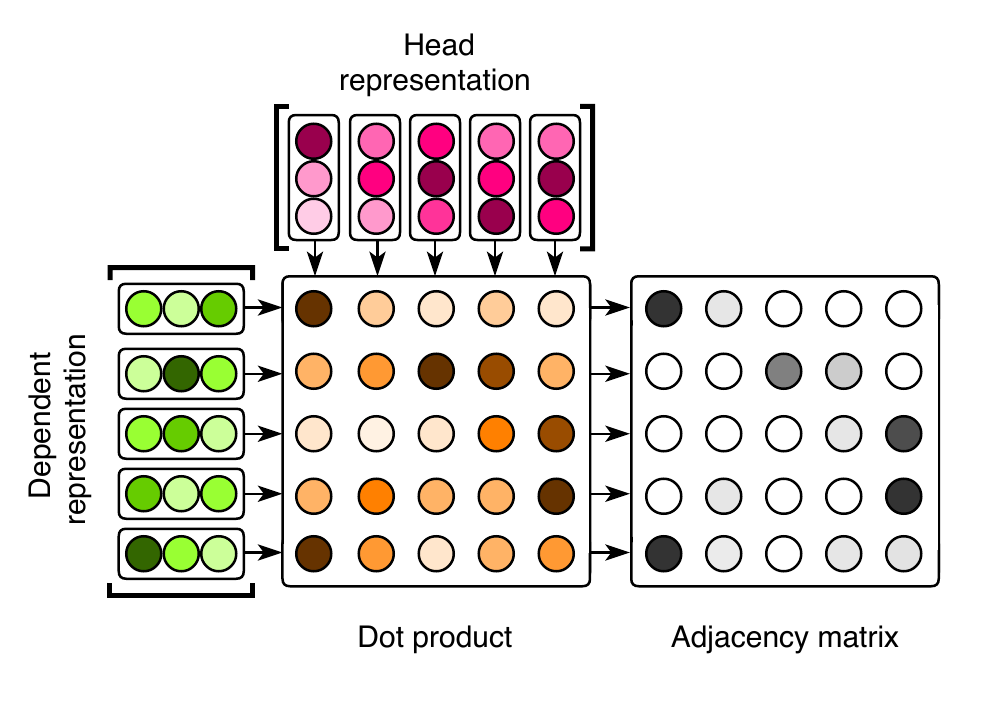}

\begin{tikzpicture}
  \SetUpEdge[lw = 0.7pt,
             color = black,
             labelcolor = white]
  \matrix (s) [matrix of nodes, column sep=7mm,
  nodes={anchor=west, text height=5pt}] {
 \sc{root\strut} & \emph{The\strut} & \emph{car\strut} & \emph{is\strut} & \emph{red\strut} \\
};
  \tikzset{EdgeStyle/.style={>=latex,->,in=90,out=90}}
  \Edge[label=\emph{cop}]([xshift=-1mm]s-1-5.north)(s-1-4.north)
  \Edge[label=\emph{det}]([xshift=-1mm]s-1-3.north)([xshift=0mm]s-1-2.north)
  \tikzset{EdgeStyle/.style={>=latex,->,in=90,out=90,looseness=1.1}}
  \Edge[label=\emph{nsubj}]([xshift=-1mm]s-1-5.north)([xshift=0mm]s-1-3.north)
  \tikzset{EdgeStyle/.style={>=latex,->,in=90,out=90,looseness=.8}}
  \Edge[label=\emph{root}]([xshift=0mm]s-1-1.north)([xshift=0mm]s-1-5.north)
  \end{tikzpicture}
\caption{\label{fig:tree}The~adjacency matrix and the~extracted dependency tree of the~sentence \textit{The car is red}.}
\end{center}
\end{figure}
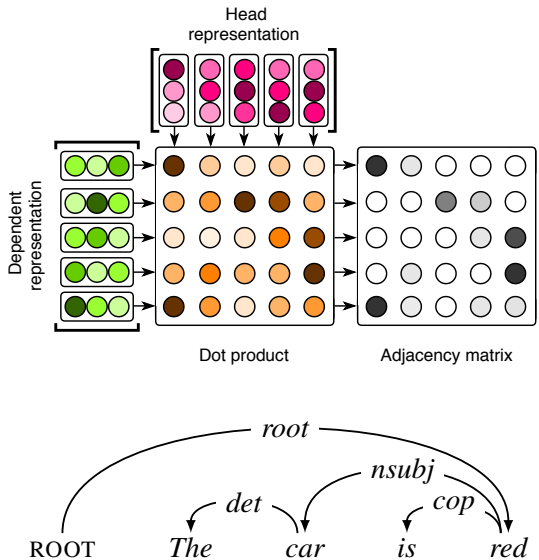

Two single fully connected layers transform features extracted by the~biLSTM encoder into head and dependent vector representations. A~fully connected dependency graph is defined with an~adjacency matrix. The~columns of the~matrix correspond to heads represented with heads' vector representations and the~rows correspond to dependents represented with dependents' vector representations. The~elements of the~adjacency matrix, in turn, are dot products of all pairs of the~head and dependent vector representations. Soft-max function is then applied to each row of the~matrix to predict the~adjacent head-dependent pairs (see Figure \ref{fig:tree}).

\subsubsection{Loss Function}
\label{sec:loss}
In order to force the~network to predict the~correct head for each dependent and thus a~correct dependency tree, the~cross-entropy loss function is used for each row in the~adjacency matrix. Note however that such formulation of the~problem can lead the~network to predict an~adjacency matrix with cycles.

We aim to get an~adjacency matrix for which a~simple greedy algorithm would suffice to construct the~correct tree. Therefore, we propose an~additional `cycle-penalty' loss function which reduces the number of cycles in the~predicted adjacency matrix:

$$\mathrm{loss}(A) = \sum^{K}_{k=1} \mathrm{tr}(A^k)$$

The non-zero trace of $A^k$ indicates that there are the~paths of the~length $k$ in the~graph represented by the~adjacency matrix $A$ \cite{cycles:1965}. Therefore, by minimizing the~sum of the~traces of the~subsequent powers of $A$ we reduce the~number of cycles in the~predicted graph. In an~ideal scenario $K$ should be equal to the~length of the~sentence, but in practice even $K=3$ helps to reduce the~number of cycles. The final loss used to train the~arc prediction model is a sum of cross-entropy loss and `cycle-penalty' loss.

If the~smoothed adjacent matrix still contains cycles, Chu-Liu-Edmonds algorithm \cite{chuLiu:65,Edmonds:67} is applied to extract the~properly built dependency tree in the~final step of the~prediction procedure.

\subsubsection{Label Prediction}
In order to predict the~label for each arc of the~predicted dependency tree, the~vector representations of the~arc's head and its dependent are calculated. These representations do not correspond to those used during the~arc prediction, but they are obtained in a~similar way. 
The~estimated vector of the~dependent is concatenated with the~weighted average of its predicted head vector. The~weights correspond to probabilities of a~word being the~dependent's head predicted by the~arc model described in the~previous section. 
It is not possible to take just the~vector of a~single predicted head, because it would prevent the model to be trained together with the rest of the system, as argmax operation is not differentiable.
The~concatenated vector representations are then fed to a~single fully connected layer with soft-max activation function.

\section{Implementation Details}
\label{sec:implementationDetails}
\subsection{Network Hyperparameters}
\label{sec:hyperparameters}
\paragraph*{Word Embedding} We use 300-dimensional fastText word embeddings \cite{grave2018learning},\footnote{\url{https://github.com/facebookresearch/fastText/blob/master/docs/crawl-vectors.md}} which are then converted to 100-dimensional vectors by a~single fully connected layer. The~embedding is not available for some languages, i.e. for Old Church Slavonic (`cu\_proiel'), Old French (`fro\_srcmf'), Gothic (`got\_proiel'), Kurmanji (`kmr\_mg'), North Sámi (`sme\_giella') or it seems incorrect, i.e. in Slovak (`sk\_snk'). Therefore, we estimate the~embedding for these languages during the~training of the~whole system.
\paragraph{Character Embedding} The~character level embedding is calculated with three convolutional layers with 512, 128 and 64 filters with dilation rates equal to 1, 2 and 4. All of the filters have the~kernel of size~3. The~input character embedding has the~size of 64.

\paragraph*{Final Word Embedding} The~final word embedding is the~concatenation of the~100-dimensional word embedding and the~64-dimensional character-based word embedding. It has thus 164 dimensions.

\paragraph*{Feature Extraction} Two biLSTM layers with 512 hidden units are used to extract the~final features.

\paragraph*{Tagger} The tagger uses a~fully connected network with the~hidden layer of the~size 64. The~model to predict morphological features uses the~hidden layer of 128 neurons.

\paragraph*{Lemmatizer} The lemmatizer uses three convolutional layers with 256 filters and dilation rates equal to 1, 2 and 4. All of the filters have the kernel of size~3. Then the~final convolutional layer with the~kernel size equal to 1 is used to predict lemmas. The~input characters, represented as the~embeddings with 256 dimensions, are concatenated with the~features extracted with the~biLSTM encoder and reduced to 32 dimensions with a~single fully connected layer. 

\paragraph*{Parser} The arc model uses heads' and dependents' vector representations with 512 dimensions. The~labelling model uses 128-dimensional vectors.

All fully connected layers use \textit{tanh} activation function and all convolutional layers use rectified linear unit \cite[ReLU,][]{relu:2010}.

\subsection{Regularization}
We apply both Gaussian Dropout (with the~dropout rate of 0.25) and Gaussian Noise (with the~standard deviation on 0.2) to the~final word embedding\footnote{\url{https://keras.io/layers/noise/}} 
and after processing each biLSTM layer.
All fully connected layers use the~standard dropout \cite{dropout} with the~dropout rate of 0.25. The~biLSTM layers use both the~standard and recurrent dropout with the~rate of 0.25.
Moreover, the biLSTM and convolutional layers use L2 regularization with the~rate of $1 \times 10^{-6}$ and the~trainable embeddings use L2 regularization with the~rate of $1 \times 10^{-5}$.

\subsection{Training}
We use cross-entropy loss for all parts of the system. The loss for the~arc prediction model is a sum of cross-entropy loss and novel loss (see Section \ref{sec:loss}). The~final loss is the~weighted sum of losses with the~following weights for each task:
\begin{itemize}
    \item 0.05 for part-of-speech tagging,
    \item 0.2 for morphological features prediction,
    \item 0.05 for lemmatization,
    \item 0.2 for arc prediction,
    \item 0.8 for label prediction.
\end{itemize}

The~whole system is optimized with ADAM \cite{Kingma:2014} with the~learning rate equal to $0.002$ and $\beta_1 = \beta_2 = 0.9$. Typically, the~batch size of approximately 2500 words is used, however for a~few of the~smallest treebanks the~batch size is reduced to 1000 or even 75 words. Each batch consists of sentences with a~similar length, but the~ordering of batches is randomized within each epoch. Each observation (i.e. sentence) is weighted with the~log of the~sentence length that forces the~model to focus on longer (and usually more difficult) sentences. The~model is trained for maximum of 400 epochs and the~learning rate is reduced twice by the~factor of two when the~validation score reaches plateau. 
For languages with multiple treebanks, first a~general model is trained on all sentences from these treebanks and then the~model is fine-tuned for each treebank.

\subsection{Self-training}
\label{sec:selflearning}
For 20 arbitrarily selected treebanks, mostly the~smallest ones, self-training method \cite{self} is used to increase the~performance of the~system. First the~model is trained in a~standard way, as described in the~previous sections. Then the~`semi-supervised' training set is built. It contains sentences with the~total of approximately 25M words taken from raw data\footnote{\url{https://lindat.mff.cuni.cz/repository/xmlui/handle/11234/1-1989}} provided by CoNLL 2018 organizers. For Uyghur language only 3M words are available. The provided data sets come either from Wikipedia or Commom Crawl. Where it is possible we choose the sentences from Common Crawl, due to longer (on average) sentence sizes.
The~pre-trained model is then used to predict dependency trees, lemmas and part-of-speech tags for all sentences in the~`semi-supervised' training set. 
Finally, the~new model is trained on this `semi-supervised' training set for only one epoch and fine-tuned on the~gold-standard training data, using the~standard training procedure.

\subsection{Languages with No Resources}
\label{sec:noresources}
Our solution for processing treebanks with no training data is very simple. We choose another language for which training data is available and train the~model on this data. The~estimated model is used for predictions in the~language with no training data. We use the~following treebank pairs:
\begin{itemize}
    \item `br\_keb' (Breton)\footnote{The first language in each row has no training data and is parsed with the~model estimated for the~second language in the~pair.} -- `ga\_idt' (Irish),
    \item `fo\_oft' (Faroese) -- `no\_nynorsk' (Norwegian),
    \item `pcm\_nsc' (Naija) -- `en\_ewt' (English),
    \item `th\_pud' (Thai) -- `vi\_vtb' (Vietnamese).
\end{itemize}

The~parallel UD treebanks for Czech
, English
, Finish
, and Swedish
, and the~treebank for modern Japanese 
are processed with the~models estimated on other treebanks for the~respective languages:
\begin{itemize}
    \item `cs\_pud' -- `cs\_pdt' (Czech),
    \item `en\_pud' -- `en\_ewt' (English),
    \item `fi\_pud' -- `fi\_tdt' (Finish),
    \item `sv\_pud' -- `sv\_talbanken' (Swedish),
    \item `ja\_modern' -- `ja\_gsd' (Japanese).
\end{itemize}

\section{Results}
\label{sec:resultOverview}
\subsection{Overview}
In the official evaluation\footnote{\url{http://universaldependencies.org/conll18/results.html}} 
\cite{udst:overview} our system ranks 3th/4th for all three main metrics (ex aequo with LATTICE and UDPipe Future for LAS). It performs particularly well on small treebanks with no development data, but a~reasonable size of the~training set. 
%
%
For example, the~system ranks 1st in terms of all three measures on Russian `ru\_taiga' treebank, 1st (MLAS and BLEX) and 2nd (LAS) on Latin `la\_perseus' treebank and spoken Slovenian `sl\_sst' treebank, and 1st (MLAS and BLEX) and 3rd (LAS) on spoken Norwegian `no\_nynorsklia' treebank.  
It is worth noting that overall MLAS and BLEX scores obtained by our system trained on small treebanks are currently the~state of the art (see Table \ref{tab:1}). With respect to LAS score, our system ranks 3rd.

\begin{table}[H]
\renewcommand\tabcolsep{8.5pt}
\begin{center}
\begin{tabular}{|l|r|r|r|}
\hline 
\bf Category & \bf LAS & \bf MLAS & \bf BLEX \\
\hline
All & 73.02 & 60.25 & 64.44\\
Big & 81.72 & 70.30 & 74.42 \\
PUD & 72.18 & 58.07 & 60.97 \\
Small & 66.90 & \bf 49.24 & \bf 54.89 \\
Low-resource & 19.26 & 1.89 & 6.17 \\
\hline
\end{tabular}
\end{center}
\caption{\label{tab:1} Official results of our system in CoNLL shared task. State-of-the-art results are in bold.}
\end{table}

Regarding to processing big treebanks, our system performs very well on Czech `cs\_fictree' treebank and English `en\_gum' treebank (1st place in MLAS and BLEX, and 2nd place in LAS), and Latin `la\_ittb' treebank (1st place in MLAS and BLEX, and 3rd place in LAS). It is very important to note that most of the~mentioned languages, i.e. Russian, Latin, Slovenian, Norwegian, and Czech, are Indo-European languages (fusional). Furthermore, for other fusional languages, e.g. Galician, Ancient Greek, Polish, Ukrainian, Dutch, Swedish, French, Italian, Spanish, Basque, our system provides quite satisfying results as well. It follows that our system is especially appropriate for processing fusional languages.

Our last observation concerns the~usefulness of external word embeddings for NLP system with a~neural architecture. The~languages without external word embeddings (see Section \ref{sec:hyperparameters}) are processed by our system significantly below its~overall performance. Hence, the~external word embeddings are crucial for a~neural NLP system.

\subsection{Impact of Loss Function}
For 15 arbitrarily selected treebanks we train the~models without the~additional loss function and we compare UAS scores of these models with UAS scores obtained by the~models estimated with the~additional loss function (with $K=3$, see Section \ref{sec:loss}). Moreover for each treebank we calculate what would be the~fraction of trees with cycles if we use the~greedy algorithm to construct the~predicted trees. 

Note that the~following results cannot be directly compared to the~official test results. First we report the~scores on the~validation set. Second we use the~gold-standard segmentation instead of the~segmentation predicted by the~baseline model.

\begin{table}[h]
\renewcommand\tabcolsep{5pt}
\begin{center}
\begin{tabular}{|l|r|r|r|r|}
\hline 
\bf Treebank & \multicolumn{2}{c|}{\textbf{UAS}} & \multicolumn{2}{c|}{\textbf{\% Cycles}} \\

\cline{2-5}

 & \bf without & \bf with & \bf without & \bf with \\ 
\hline
ar\_padt & 86.23 & 86.39 & 7.70 & 4.51 \\
bg\_btb & 92.32 & 92.45 & 1.26 & 1.52 \\
cu\_proiel & 86.94 & 86.49 & 4.19 & 3.91 \\
da\_ddt & 86.61 & 86.24 & 5.14 & 4.61 \\
de\_gsd & 87.74 & 87.64 & 3.50 & 2.63 \\
es\_ancora & 92.39 & 92.49 & 3.08 & 3.39 \\
fa\_seraji & 90.30 & 90.44 & 5.18 & 4.51 \\
got\_proiel & 83.88 & 83.57 & 5.48 & 4.57 \\
hr\_set & 90.34 & 90.50 & 6.71 & 4.95 \\
hu\_szeged & 81.99 & 82.48 & 11.34 & 9.75 \\
id\_gsd & 84.33 & 84.47 & 6.98 & 6.98 \\
lv\_lvtb & 85.69 & 85.58 & 6.28 & 4.66 \\
pt\_bosque & 92.65 & 92.63 & 1.25 & 1.43 \\
ro\_rrt & 91.04 & 90.89 & 2.39 & 2.53 \\
vi\_vtb & 68.92 & 69.02 & 15.00 & 12.63 \\
\hline
\bf Average & 86.76 & 86.75 & 5.70 & 4.84 \\
\hline
\end{tabular}
\end{center}
\caption{\label{tab:4} Comparison of the~models trained with and without the~additional loss function.}
\end{table}

The~additional loss only slightly decreases UAS (see the~second and the~third column in Table \ref{tab:4}). However, it also has only a~small impact on the~cycles reduction (see the~fourth and the~fifth column in Table \ref{tab:4}). If there is a~lot of cycles in the~graphs predicted without the~additional loss, e.g. 7.7\% cycles in `ar\_padt' (Arabic), the~number of cycles is significantly reduced with the~additional loss function, i.e. the~reduction by 3.2 p.p. If the~rate of cycles is lower, e.g. 4.19\% in `cu\_proiel' (Old Church Slavonic), fewer cycles are corrected, i.e. the~reduction by 0.28 p.p. Finally, there are four treebanks -- `bg\_btb' (Bulgarian), `es\_ancora' (Spanish), `pt\_bosque' (Portuguese), and `ro\_rrt' (Romanian), for which the~additional loss function slightly increases the~number of cycles.

\subsection{Impact of Self-training}
\label{sec:results1}
We test the~impact of self-training method on the~performance of the~system trained on 20 selected treebanks. 
Again the~models are tested on the~validation set with the~gold-standard segmentation.

\begin{table*}[h]
\renewcommand\tabcolsep{9pt}
\begin{center}

  \begin{tabular}{|l|r|r|r|r|r|r|}
  \hline 
  \bf Treebank & \multicolumn{2}{c|}{\textbf{LAS}} & \multicolumn{2}{c|}{\textbf{MLAS}} & \multicolumn{2}{c|}{\textbf{BLEX}} \\

  \cline{2-7}

   & \bf std & \bf self & \bf std & \bf self & \bf std & \bf self \\ 
  \hline
  bg\_btb & 88.84 & 89.23 & 79.39 & 80.12 & 78.89 & 79.21 \\
  da\_ddt & 83.24 & 84.89 & 72.40 & 75.76 & 76.73 & 78.93 \\
  el\_gdt & 87.89 & 89.19 & 73.55 & 76.98 & 74.06 & 77.30 \\
  eu\_bdt & 81.58 & 82.85 & 65.94 & 68.15 & 76.29 & 77.49 \\
  fa\_seraji & 86.91 & 87.22 & 80.19 & 81.18 & 80.86 & 81.36 \\
  ga\_idt & N/A & N/A & N/A & N/A & N/A & N/A \\
  he\_htb & 84.39 & 85.60 & 71.31 & 73.63 & 73.87 & 75.34 \\
  hr\_set & 86.06 & 86.22 & 71.17 & 71.61 & 79.01 & 79.38 \\
  hu\_szeged & 77.39 & 80.55 & 60.77 & 67.28 & 70.55 & 74.41 \\
  id\_gsd & 77.62 & 77.97 & 64.27 & 66.22 & 74.03 & 74.62 \\
  kk\_ktb & N/A & N/A & N/A & N/A & N/A & N/A \\
  lv\_lvtb & 80.52 & 82.67 & 65.53 & 69.11 & 71.71 & 74.19 \\
  ro\_rrt & 85.88 & 86.68 & 76.49 & 77.57 & 79.47 & 80.39 \\
  sk\_snk & 83.44 & 85.52 & 56.05 & 66.69 & 72.77 & 77.13 \\
  tr\_imst & 64.07 & 64.95 & 49.26 & 51.97 & 57.15 & 58.87 \\
  ug\_udt & 63.89 & 65.50 & 38.21 & 41.83 & 51.48 & 53.89 \\
  uk\_iu & 85.83 & 87.91 & 68.35 & 73.69 & 78.35 & 81.65 \\
  ur\_udtb & 80.91 & 81.34 & 52.92 & 53.48 & 71.09 & 71.72 \\
  vi\_vtb & 58.82 & 60.52 & 49.29 & 51.87 & 54.85 & 56.93 \\
  zh\_gsd & 77.09 & 76.71 & 65.17 & 64.88 & 69.68 & 68.93 \\
  \hline
  \bf Average & 79.69 & 80.86 & 64.46 & 67.33 & 71.71 & 73.43 \\
  \hline
  \end{tabular}

\end{center}
\caption{\label{tab:3} Comparison of the~standard (std) and self-training (self) models on the~validation set using the~gold-standard segmentation. Note that `kk\_ktb' (Kazakh) and `ga\_idt' (modern Irish) treebanks do not have validation sets, so we are unable to report any results.}
\end{table*}

Comparing the~results of the~models estimated on training data with the~results of the~models estimated with the~self-training method (see Table \ref{tab:3}), we notice that self-training significantly increases the~performance of the~system. There is an~increase for all metrics for all treebanks except for `zh\_gsd' (Chinese). On average there is an~increase of 1.2 p.p. for LAS, 2.9 p.p. for MLAS and 1.7 p.p. for BLEX.

\section{Conclusion}
\label{sec:conclusions}

We described the ICS PAS system which took part in CoNLL 2018 shared task. Our goal was to build one system for preprocessing natural languages, i.e. for part-of-speech tagging, lemmatisation and dependency parsing. The~three system's modules -- tagger, lemmatizer and parser -- are jointly trained. The~proposed neural system ranks 3th/4th in the~official evaluation of the~shared task. It is worth nothing that the~system is especially useful for estimating the~models on relative sparse data (small treebanks), as it overcame other systems in terms of MLAS and BLEX. Furthermore, our system is especially appropriate for processing Indo-European fusional languages.

The~self-training procedure significantly increases the~performance of the system. The proposed loss function, in turn, has only a~slight impact on the~cycles reduction and UAS scores. The~external word embeddings are crucial for our neural-based system.

\vspace*{1cm}
\section*{Acknowledgements}
The research presented in this paper was founded by SONATA 8 grant no 2014/15/D/HS2/03486 from the National Science Centre Poland and by the Polish Ministry of Science and Higher Education as part of the investment in the CLARIN-PL research infrastructure.

\bibliography{udst2018}

\bibliographystyle{acl_natbib}

\end{document}